\documentclass[conference]{IEEEtran}
\IEEEoverridecommandlockouts
\usepackage{stfloats}

\usepackage{cite}
\usepackage{amsmath,amssymb,amsfonts}
\usepackage{algorithmic}
\usepackage{graphicx}
\usepackage{textcomp}
\usepackage{xcolor}

\usepackage{subfigure}
\usepackage{float} 

\def\BibTeX{{\rm B\kern-.05em{\sc i\kern-.025em b}\kern-.08em
    T\kern-.1667em\lower.7ex\hbox{E}\kern-.125emX}}

\newcommand{\comment}[1]{}

\newcommand{\BEA}{\begin{eqnarray}}
\newcommand{\EEA}{\end{eqnarray}}

\newcommand{\bq}{\begin{equation}}
\newcommand{\eq}{\end{equation}}
\newcommand{\be}{\begin{eqnarray}}
\newcommand{\ee}{\end{eqnarray}}
\newcommand{\ba}{\begin{align}}
\newcommand{\ea}{\end{align}}

\begin{document}

\title{Nonnegative matrix factorization and the principle of the common cause \\
%\thanks{The work was supported by HESC of Armenia, grants No 21AG-1C038 and No 24FP-1F030}
}

\author{\IEEEauthorblockN{Edvard Khalafyan}
\IEEEauthorblockA{
%\textit{Department of Applied Mathematics and Computer Science} \\
\textit{Moscow Institute of Physics and Technology}\\
Moscow, Russia \\
edvardkhalafyan@gmail.com}
\and
\IEEEauthorblockN{%2\textsuperscript{nd} 
Armen Allahverdyan}
\IEEEauthorblockA{
%\textit{Department of Quantum Technologies} \\
\textit{Alikhanyan National Laboratory}\\
Yerevan, Armenia \\
a.allahverdyan@aanl.am}
\and
\IEEEauthorblockN{Arshak Hovhannisyan}
\IEEEauthorblockA{
%\textit{Department of Quantum Technologies} \\
\textit{Alikhanyan National Laboratory}\\
Yerevan, Armenia \\
arshak.hovhannisyan@aanl.am
}
}

\maketitle

\begin{abstract}
Nonnegative matrix factorization (NMF) is a known unsupervised data-reduction method. The principle of the common cause (PCC) is a basic methodological approach in probabilistic causality, which seeks an independent mixture model for the joint probability of two dependent random variables. It turns out that these two concepts are closely related. This relationship is explored reciprocally for several datasets of gray-scale images, which are conveniently mapped into probability models. On one hand, PCC provides a predictability tool that leads to a robust estimation of the effective rank of NMF. Unlike other estimates (e.g., those based on the Bayesian Information Criteria), our estimate of the rank is stable against weak noise. We show that NMF implemented around this rank produces features (basis images) that are also stable against noise and against seeds of local optimization, thereby effectively resolving the NMF 
nonidentifiability problem. 
On the other hand, NMF provides an interesting possibility of implementing PCC in an approximate way, where larger and positively correlated joint probabilities tend to be explained better via the independent mixture model. We work out a clustering method, where data points with the same common cause are grouped into the same cluster. We also show how NMF can be employed for data denoising.

%TLDR: Two important concepts - Nonnegative matrix factorization and the principle of the common cause -
%and B are related to each other and each solves the other's open problem.

\end{abstract}


\section{Introduction}

Nonnegative matrix factorization (NMF) was proposed and developed in data science \cite{finn,lee,gillis-review}. As compared to other matrix factorization methods (e.g., PCA), NMF requires that positive data matrices be factored into positive matrices, which makes NMF outcomes interpretable as features and their weights. 

Matrix factorizations are also (noisy) data compression methods, because the initial data matrix is represented via a smaller number of parameters. In the sense of the Frobenius distance, the optimal compression method of the original matrix to a matrix of a given rank is known to be PCA (principal component analysis, which, for the present purpose, is taken to be the same as the truncated SVD). 
PCA achieves the global error minimization in a polynomial number of steps. In contrast to PCA, NMF 
generally achieves only a local error minimization \cite{gillis-review}. The existing algorithms for local optimization do not provide any guaranteed distance from the global optimum, which is algorithmically hard to reach \cite{complexity-vava}. Moreover, the global error minimum is not unique, since the exact NMF is generally 
nonidentifiable \cite{gillis-review}. As a result, when mentioning interpretable features of NMF obtained via local error minimization, it is often unclear what these features are and why they are useful. 
Our discussion leads to several open problems on the structure and function of NMF.  

{\it (i)} How do the extracted features depend on NMF contingencies, such as noise in data or the seed for local optimization? How does this relate to the nonidentifiability of NMF?

{\it (ii)} How to set the NMF hyperparameters, the most important one being the dimension of the factorization (effective rank)? This problem attracted much attention for NMF \cite{fogel,mdl}, and 
is closely related to the analogous problem for PCA.

{\it (iii)} In what sense do features extracted from data explain this data? Do they apply for the feature-based clusterization? 

There is another field, where the idea of exact NMF emerged much earlier: probabilistic causality \cite{reich,suppes}. Here the dependence between two random variables is explained
via a third random variable (common cause) which makes them conditionally independent. The ensuing principle of the common cause (PCC) \cite{reich} was later generalized to the Causal Markov condition and is foundational for the modern probabilistic causality \cite{janzing}. Beyond that field, PCC found important application in several other disciplines: quantum mechanics \cite{wharton} (Bell's inequalities, (non)locality), statistical physics (direction of time) \cite{penrose1962direction}, cognitive psychology \cite{rehder2003causal}, evolutionary biology \cite{bio,sober_common}, decision-making \cite{simpson}, {\it etc}.
PCC explains probabilistic dependencies via conditional independence. This leads to an important open question.

{\it (iv)} What is the relationship between NMF and PCC?

Here we answer the open questions and show that they are interrelated. We work with gray-scale image databases, which are conveniently mapped to probabilistic models. They ensure the applicability of both NMF and PCC. 

One difference between NMF and PCC is that NMF is implemented in an approximate form, while PCC is regarded as an exact relation. We show that this difference is constructive both for NMF and PCC. On one hand, it allows generalizing PCC by requiring that relatively large and positively correlated probabilities are explained better. In addition, the entropy increases during approximation, even for a single image; cf.~{\it (iv)}. On the other hand, relations deduced from the exact PCC are useful for estimating the effective rank; see {\it (ii)}. The estimate is stable against (weak) noise, in contrast to the estimators proposed in the literature, e.g., those based on the Bayesian Information Criteria (BIC).

For answering {\it (i)} and {\it (iii)}, we show that features (basis images) extracted from NMF are stable against noise and the seed of local optimization. The stability relates to a specific range of NMF ranks around the above mentioned estimate. Stable and interpretable features (basis images) can be employed for clustering the data, i.e., gathering all data items that have this feature with a large probability. Such a clusterization principle is known in biology \cite{sober_common}: two items are put in one cluster, if they have a common cause \cite{rehder2003causal}.

To understand the structure of NMF for gray-scale image datasets, we applied it to the denoising problem. 
The denoising here is based on the prior information that comes from the fact that the image belongs to a specific dataset (dictionary-based denoising \cite{review_denoising}). This task does not require interpretability and allows a direct comparison in terms of accuracy between various matrix factorization methods. 

The paper is organized as follows. Section \ref{NMF} reviews NMF and PCC and discusses previous work. Datasets and their manipulation methods are described in section \ref{data}. Section \ref{Restimation} estimates the effective rank of NMF via ideas of efficient probabilistic prediction inherited from PCC. Section \ref{basistability} discusses the stability of basis images (the outcomes of NMF for image datasets) under noise and binarization. Section \ref{appro} discusses what type of approximation is suggested by NMF to PCC. Section \ref{nato} discusses the concept of natural clusterization, which is achieved via NMF and PCC.
Denoising is discussed in section \ref{denoise}. We summarize in the last section.

\section{Previous work: NMF and PCC}
\label{NMF}

\subsection{NMF and its Optimization Metrics}

NMF approximates a nonnegative data matrix $P$ as the product of two nonnegative matrices:
\begin{align}
\label{a1}
&{P}_{\pi i} \approx \hat{P}_{\pi i}\equiv {\sum}_{b=1}^{{R}} B_{\pi b} W_{bi},
\qquad \pi=1,..,N,\\ 
& {P}_{\pi i}\geq 0,~ B_{\pi b}\geq 0,~  W_{bi}\geq 0, \qquad
i=1,...,M,
\label{x1}
\end{align} 
where all matrices in (\ref{a1}) have nonnegative elements, and where the approximation in (\ref{a1}) is frequently understood via a locally minimal Frobenius distance $||P-\hat P||_2$:
\be
\label{fro}
\hat P={\rm argmin}_{B,W}\Big[||P-\hat P||^2_2\,;\, B_{\pi b}\geq 0,~W_{bi}\geq 0\Big],\\
||P-\hat P||^2_2
={\sum}_{\pi=1\, i=1}^{NM}\Big[ {P}_{\pi i} - \hat{P}_{\pi i} \Big]^2.
\label{fropo}
\ee
The approximation in (\ref{a1}) relates to $R$ being smaller than the (positive) rank of $P$; see section \ref{nonident}. Eq.~(\ref{a1}) achieves (noisy) compression when $NM$ elements of $P_{\pi i}$ are approximately represented by a smaller number of parameters:
\be
\label{compr}
R(N+M)<NM.
\ee
We can employ (instead of (\ref{fropo})) the Kullback-Leibler (KL) divergence in 
approximation (\ref{a1}) \cite{kl}:
\be
{\rm KL}[P,\hat P]
={\sum}_{\pi=1\, i=1}^{NM} \Big[ 
P_{\pi i}\ln[{P_{\pi i}}/{\hat P_{\pi i}}]-P_{\pi i}+\hat P_{\pi i}
\Big].
\label{ku}
\ee
The advantage of using ${\rm KL}[P,\hat P]$ instead of $||P-\hat P||_2$ in (\ref{fro}) is that when $\hat P$ is determined from a local minimum of ${\rm KL}[P,\hat P]$, it conserves partial 
normalization \cite{kl}:
\be
{\sum}_{\pi=1}^N {P}_{\pi i} = {\sum}_{\pi=1}^N \hat{P}_{\pi i},\quad
{\sum}_{i=1}^M {P}_{\pi i} = {\sum}_{i=1}^M \hat{P}_{\pi i}.
\label{norma}
\ee
Eqs.~(\ref{norma}) are rigorous features of local minimas of (\ref{ku}). Minimizing the Frobenius distance $||P-\hat P||_2$ satisfies (\ref{norma}) approximately, with a precision much larger than in (\ref{a1}). Hence we can use the Frobenius distance and employ (\ref{norma}) without serious numerical errors. To achieve local minimization in (\ref{fro}), we employed the known alternating least-squares algorithm proposed in \cite{finn} and implemented in the scikit-learn \cite{scikit}.

\comment{
where the local minimization is done sequentially: first $||P-BW||_2$ is minimized over $B$ for a given $W$, the resulting $B$ is held fixed, and the minimization is done over $W$, and so on \cite{gillis-review}. Our results were checked with the Kullback-Leibler divergence (\ref{ku}), and sometimes we report both. }

\subsection{Probabilities and gray-scale images}

We focus on two concrete examples of (\ref{a1}). For the first example, $\{\pi \}_{\pi=1}^N$ and $\{i \}_{i=1}^M$ are realizations of two random variables $\Pi$ and ${\cal I}$, while $P_{\pi i}=p(\pi, i)$ refers to their joint probability. Now $\{b\}_{b=1}^R$ are realizations of the third random variable ${\cal B}$ which makes this joint probability approximately conditionally independent with a conditional probability $B_{\pi b}=p(\pi| b)$ and joint probability $W_{bi}=p(b,i)$. 

For the second example, $P_{\pi i}$ with $255\geq P_{\pi i}\geq 0$ is the gray-scale intensity of the pixel $\pi$ in image $i$. The index $b$ in (\ref{a1}) refers to basis images, $B_{\pi b}$ is the intensity of the pixel $\pi$ within the basis image $b$, and $W_{b i}$ is the weight of $b$ in image $i$. The second example is reduced to the first one: note the freedom of choosing parameters $\kappa_b>0$ that leave (\ref{a1}) intact: 
\be
B_{\pi b}\to B_{\pi b} \kappa_b, \qquad W_{bi}\to W_{bi}/\kappa_{b}. 
\label{a2}
\ee
Using (\ref{norma}, \ref{a2}), we write (\ref{a1}) as
\begin{align}
\label{a3}
& {p}(\pi, i)
\approx 
\hat{p}(\pi, i)
\equiv {\sum}_{b=1}^{{ R}} p(\pi|b) p(b,i),\\ 
&p(\pi,i)\equiv {P}_{\pi i}/{\cal P},\quad 
{\cal P}\equiv {\sum}_{\pi=1\, i=1}^{NM}P_{\pi i}, \\ 
& p(\pi|b)\equiv\frac{B_{\pi b}}{\sum_{\rho=1}^N B_{\rho b}  }, \quad
p(b,i)\equiv\frac{1}{ {\cal P} }  W_{bi}{\sum}_{\rho=1}^N B_{\rho b} 
\label{a303}
\end{align} 
where $p(\pi|b)$ is the conditional probability. Note that (\ref{a303}) is invariant under (\ref{a2}).
Other probabilities are calculated from (\ref{a3}, \ref{a303}); e.g. $p(\pi|i)$ is the relative intensity of the pixel $\pi$ in image $i$, where the relative means that it is normalized over (compared with) all other pixels within the same image $i$. $p(\pi|b)$ is the same quantity, but within basis image $b$. $p(b|i)$ is the weight by which image $i$ contains basis image $b$. 

\subsection{Positive rank and nonidentifiability}
\label{nonident}

The positive rank ${\rm rank}_+[\{ P_{\pi i}  \}]={\rm rank}_+[\{{p}(\pi, i)\}]$ is the minimal $R$ for which (\ref{a1}, \ref{a3}) are exact equalities holding (\ref{x1}):
\be
\label{a33}
{p}(\pi, i) = {\sum}_{b=1}^{{ R}} p(\pi|b) p(b,i)= {\sum}_{b=1}^{{ R}}p(b) p(\pi|b) p(i|b).
\ee
The positive rank satisfies \cite{rank-nonnegative}:
\be
\label{nicomach}
{\rm rank}[\{{p}(\pi, i)\}]\leq {\rm rank}_+[\{{p}(\pi, i)\}]\leq {\rm min}[N,M].
\ee
The first inequality in (\ref{nicomach}) arises because for each $b$, $\{p(\pi|b) p(b,i)\}_{\pi i}$ are rank-1 matrices \cite{rank-nonnegative}. The second inequality follows from the fact that in (\ref{a33}) we can take ${\cal B}={\cal I}$ or ${\cal B}={\Pi}$. For full rank matrices ${\rm rank}[\{{p}(\pi, i)\}]={\rm min}[N,M]$, and we do not get from (\ref{nicomach}) compression in the sense of (\ref{compr}). Many interesting applications of (\ref{a3}) are precisely in the domain $R<{\rm rank}[\{{p}(\pi, i)\}]$, where (\ref{a3}) is approximate, but is supposed to retain certain pertinent features of $\{{p}(\pi, i)\}$. 

The exact relation (\ref{a33}) is known to be non-unique: for a {\it generic} $p(\pi,i)$, there are 
many different $p(\pi|b)$ and $p(b,i)$ that satisfy (\ref{a33}). 
Different means not related to each other via permuting $\{b\}_{b=1}^R$. Hence from the viewpoint of the mixture estimation, the problem (\ref{a33}) is not identifiable \cite{donoho,gillis-review,hovh2023}; see section \ref{dato}.

Ref.~\cite{hovh2023} proposed a generalized maximum-likelihood principle for resolving this nonidentifiability, and reviewed related approaches. Ref.~\cite{gillis-review} reviews several classes of data (i.e., matrices $p(\pi,i)$) for which (\ref{a33}) is unique at $R={\rm rank}_+[p(\pi,i)]$. However, these classes are not stable under perturbations.

\subsection{The principle of the common cause (PCC)}
\label{commi}

The random variable ${\cal B}=\{b\}_{b=1}^R$ in (\ref{a33}) is the common cause for ${ \Pi}=\{\pi\}_{\pi=1}^N$ and ${\cal I}=\{i\}_{i=1}^M$ \cite{reich,suppes}. One motivation for this definition is that the joint probability found from (\ref{a33}) is conditionally independent (screening):
\BEA
\label{screening}
p(\pi,i|b)=p(\pi|b)p(i|b),~ p(\pi|i)={\sum}_{b=1}^Rp(\pi|b)p(b|i).
\EEA
${\cal B}$ {\it explains} the dependence (correlations) between $\Pi$ and ${\cal I}$ if conditional independence is believed to be the natural state of affairs. This probabilistic explanation is similar to how separate basis images in (\ref{a1}) explain the dataset. Note that if ${\cal B}=\{b\}_{b=1}^R$ is aggregated (coarse-grained) to fewer events, those events will generally not hold the analogue of (\ref{a33}). 
Hence the exact relation (\ref{a33}) requires the minimal value of $R$ equal to ${\rm rank}_+[\{p(\pi,i)\}]$. 

PCC was proposed in a specific probabilistic context \cite{reich,suppes}, which was later generalized to the causal Markov condition, directed acyclic graphs, and the Markov blanket concept, pillars of the modern probabilistic causality \cite{janzing}. Screening condition (\ref{screening}) is behind both causal and non-causal feature selection concepts; see \cite{guyon_review, causal_non_causal_review} for reviews. PCC is a part of a general scientific principle that looks for unique explanations of diverse (but correlated) phenomena \cite{wharton,penrose1962direction,rehder2003causal,bio,sober_common,simpson}.

Despite many existing applications, there is a permanent quest for generalizing PCC \cite{mazzola2019generalised,horwich1987asymmetries,uffink,imprecise}. 
Why would the independence always be considered as the benchmark of explanation? 
Do joint probabilities of any magnitude have to be explained in the same way? Probabilistic dependence involves at least two aspects (positive and negative correlations). Is it appropriate to explain them both the same way? To some extent, these questions are answered in section \ref{appro}.

\section{Data and Methods}
\label{data}

\subsection{Description of the datasets and their NMF outcomes}
\label{dato}

We employed three different gray-scale image datasets.

{\it Swimmer} is a simple dataset of 256 images with 169=13x13 black-white pixels per image \cite{donoho}. Each image represents a backbone with four limbs; each limb can be in 4 ($4^4=256$) different positions, and the backbone is common to all images. Given the matrix $P_{\pi i}$ of the dataset, its positive rank is ${\rm rank}_+[P]=17$. Indeed, for $R=17$, (\ref{a1}) holds exactly (without approximation) with $B_{\pi b}$ ($m=1,..,17$) being the following basis images: one backbone (present in all images) and 16 images for separate positions of limbs. Now $p(i|b)=0,1$ in (\ref{a33}) are deterministic weights by which 17 basis images combine into the original dataset. This exemplifies the nonidentifiability, since the exact NMF can have different sets of basis images depending on how the backbone is distributed. 

Our results show that depending on the seed of local optimization in (\ref{fro}), the error (\ref{fropo}) of the optimization suddenly drops for $16\leq R\leq 18$. Hence, ${\rm rank}_+[P]$ can be approximately determined from analyzing the error. This feature is not stable against weak noise (defined by (\ref{noise}) below). For $R=16, 17$ the basis images found from (\ref{a1}) amount to 16 limbs with the backbone shared between them. The amount of sharing depends on the seed of the local optimization in (\ref{a1}). The partial recovery of the backbone (which is shared among all images) is not stable against a weak noise. 

{\it Olivetti} consists of 400 images with $4,096=64\cdot 64$ pixels each \cite{olivetti_faces_dataset} (we obtained Olivetti data via \cite{scikit}). It represents 40 individuals, each with 10 images taken from different perspectives. Because it contains images of the same person, it will be useful to us; see section \ref{nato}. 

{\it UTKFace} (UTK) is a well-regarded dataset in the field of facial analysis, often used for age estimation, gender classification, and ethnicity recognition \cite{zhifei2017cvpr} (we obtained UTK data via \cite{jangedoo_utkface}). It contains 23,708 images. We represent every image with $2,500=50\cdot 50$ pixels.

Note that for Swimmer and UTK (but not for Olivetti) the number of pixels in a given image is smaller than the number of database images, i.e. $N<M$ in (\ref{a1}).

\subsection{Noise and binarization}

\begin{figure}[t]
    \centering
    \includegraphics[width=1.\linewidth]{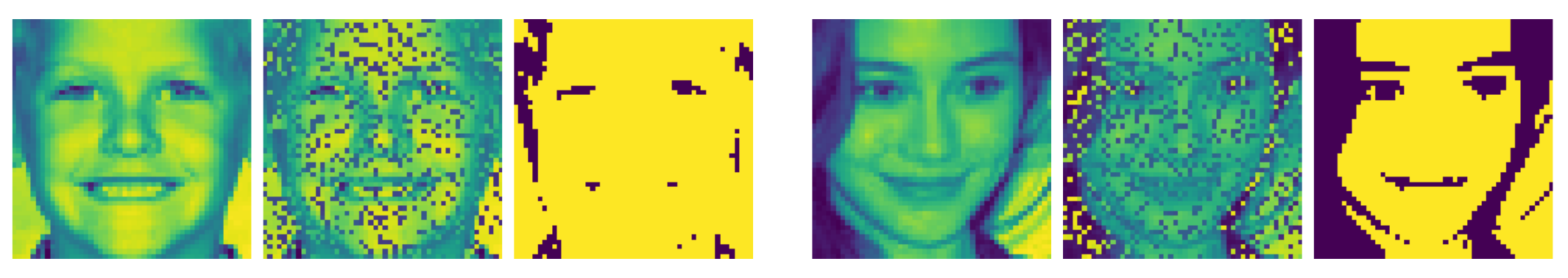}
    \caption{Two images from UTK dataset [see section \ref{dato}]: the original image (first column),  binary noise corrupted image (second column, the error probability $\xi=0.25$), the binarized image (third column).}
    \label{fig:UTK}
\end{figure}

For introducing noises and perturbations in the simplest way, we shall rescale each pixel of each dataset as follows. Originally, all pixels $P_{\pi i}$ of a given image $i$ are gray-scale: $0\leq P_{\pi i}\leq 255$. We rescale $P_{\pi i}\to P_{\pi i}/255$, $0\leq P_{\pi i}\leq 1$. Next, we need noisy versions of images. The simplest model is a symmetric noise, which acts independently on each pixel, leaving it intact with probability $1-\xi$ and flipping it: 
\begin{align}
    P_{\pi i} \to 1-P_{\pi i}, {\rm~ with~ probability}~ \xi.
    \label{noise}
\end{align}
This symmetric noise model is useful for assessing the robustness of algorithms to bit-level perturbations and simulates scenarios where random, unbiased errors occur in binary data representations.
Another variant of data perturbation is binarization. In this process, each (rescaled) pixel is compared with a fixed threshold of 0.5, and the binarized image is produced as follows: $P_{\pi i}\to 1$, if $P_{\pi i}\geq 0.5$, and $P_{\pi i}\to 0$, if $P_{\pi i}< 0.5$. This operation transforms grayscale images into binary ones, effectively simplifying the representation. Examples of various perturbation methods applied to UTK dataset are presented in Fig.~\ref{fig:UTK}, where we show original images, their noise-corrupted versions and their binarizations.

\section{Estimation of the effective rank $R$ via the common-cause principle}
\label{Restimation}

\subsection{Estimation of $R$ via predictability}
\label{predo}

\subsubsection{Clean datasets}

One meaning of the common cause is illustrated by the theorem that follows from (\ref{a33}), and was first proved in \cite{suppes}; see also below. If $p(\pi|i)$ is a predictor of pixels $\{\pi\}$ in a given image $i$, then the common cause $\{b\}$ (the set of basis images) is generally a better predictor:
\begin{align}
&\forall (\pi,i)~ \exists b_1(\pi,i)~ \exists b_2(\pi,i)~:\nonumber\\
&p(\pi|b_1(\pi,i))\leq p(\pi|i)\leq p(\pi|b_2(\pi,i)).
\label{a5}
\end{align}
Recall that $p(\pi|i)$ quantifies the statistical influence of $i$ on $\pi$; making this influence closer to 1 or to 0, improves the prediction. In that sense, (\ref{a5}) shows that $b_1(\pi,i)$ and $b_2(\pi,i)$ have more predictive power than $i$. The two inequalities in (\ref{a5}) are proved by contradiction. Assume that $p(\pi|b)> p(\pi|i)$ for all $b$. Multiply both sides of this inequality by $p(b,i)$ and assume for simplicity that $p(b,i)>0$; dropping this assumption for certain $p(b,i)$ does not change the conclusion. We get: $p(\pi|b)p(b,i)> p(\pi|i)p(b,i)$. Now use $p(\pi|b)=p(\pi|b,i)$ [follows from (\ref{screening})] and sum over all $b$. The resulting contradiction $p(\pi,i)>p(\pi,i)$ proves the first (and likewise the second) inequality in (\ref{a5}). 

Eq.~(\ref{a5}) makes intuitive sense in the context of images [see the discussion after (\ref{a303})]: for a given pixel $\pi$ in a given image $i$ there is always a basis image $b_2$, where this pixel $\pi$ is brighter. One scenario how this can happen is that the pixel belongs to a small set of connected pixels (part of the image), which came out as a localized structure in $b_2$. Likewise, there is a basis image $b_1$, where this pixel is dimmer, i.e. there are basis images, where the parts in which $\pi$ participates are absent. 

Since (\ref{a5}) follows from the exact (\ref{a33}), we can use the intuition of (\ref{a5}) for deciding on the minimal sensible value of $R$, which appears in {\it approximate} relation (\ref{a3}). Numerical experiments 
for clean datasets show that there is a minimal value $R=R_c$ such that (\ref{a5}) starts to hold for all pixels of a dataset provided that $R\geq R_c$. For $R\to R_c$ (from below), the fraction of valid inequalities in (\ref{a5}) converges to 1. Note that $R_c$ is defined in local minima of (\ref{fropo}) or (\ref{ku}). If $R_c$ is defined via exact relation (\ref{a33}), we get $R_c = {\rm rank}_+[P]$.

For all datasets we studied, we observed $R_c< {\rm rank}_+[P]$; e.g. for Swimmer $R_c=14< {\rm rank}_+[P]=17$. For Olivetti and UTK we get $R_c\ll {\rm rank}_+[P]$; see Table \ref{table:1}.

\begin{table}[htbp]
\caption{\textrm{ $R_c$ for 3 noiseless datasets: Swimmer, Olivetti, and UTK. The results were checked with 10 different seeds of the local optimization in (\ref{a1}). Changing the size (1500, 3000, 6000) of the random sample for the clean UTK does not affect the results. 
} }
\begin{center}
\begin{tabular}{|c||c|c|c| }
\hline
Dataset & Swimmer & Olivetti & UTK \\ \hline
$R_c$ & 14 & 26 & 30\\ \hline
\end{tabular}
\label{table:1}
\end{center}
\end{table}

\subsubsection{Noisy datasets}
\label{akhmuk}
Adding noise (\ref{noise}) to UTK does not change the above behavior essentially, but $R_c$ increases as expected: $R_c(\xi)>R_c(0)$; e.g., $R_c(0.05)=46>R_c(0)=30$.

A new effect appears upon adding noise to Olivetti and to Swimmer. Instead of the convergence noted above for clean Swimmer, clean Olivetti and UTK, we get a delayed regime, where there are pairs $(\pi,i)$ for which either $p(\pi|b)> p(\pi|i)$ $\forall b$, or 
$p(\pi|b)< p(\pi|i)$ $\forall b$, i.e. at least one inequality in (\ref{a5}) is invalid. The number $\tau NM$ of such pairs is relatively small: $\tau\ll 1$. Now $\tau$ slowly decreases upon increasing $R$. We called this ``outlier-pixel'' effect; it is present in Olivetti, because it is small and contains repeated images, and in Swimmer, because it is small and sparse. The outlier-pixel effect is described via $R(\xi,\tau)$, which depends on the noise intensity $\xi$ and the parameter $\tau\ll 1$. For $R>R(\xi,\tau)$ only $\tau NM$ among $NM$ inequalities (\ref{a5}) are invalid. 

Two threshold values of $\tau$ are $\tau_1=M^{-1}$ (only a single pixel in each image is invalid) and $\tau_2=(MN)^{-1}$: a single pixel in the dataset is invalid. These thresholds can be used depending on the purpose of applying NMF; e.g. whether one wants to account for almost all pixels, or neglect noisy pixels. In the latter case, we can define $R_c=R(\xi,\tau_1)$; for the former case $R_c=R(\xi,\tau_2)$. Table~\ref{table:noise} shows $R(\xi,\tau)$ for the datasets. 

\begin{table}
\caption{$R(\xi,\tau)$ for noisy Swimmer ($M=256$, $N=13^2$), Olivetti ($M=400$, $M=64^2$), and
UTK ($M=1500$, $N=50^2$); see (\ref{noise}) and section \ref{akhmuk} for definitions of (resp.) $\xi$ and $\tau$.  }
\begin{center}
\begin{tabular}{|c||c|c| }
\hline
Dataset & \multicolumn{2}{|c|}{Swimmer} \\ 
\hline
thresh.  & $\tau = 10^{-2}$ & $\tau = 10^{-3}$   \\ \hline 
$\xi = 0.05$  & 44 & 76   \\ \hline 
$\xi = 0.15$  & 34 & 50    \\ \hline 
$\xi = 0.25$  & 22 & 30    \\ \hline \hline 
Dataset & \multicolumn{2}{|c|}{Olivetti} \\ \hline 
thresh.  & $\tau = 10^{-4}$ & $\tau = 10^{-5}$   \\ \hline 
$\xi = 0.05$  & 27 & 43   \\ \hline \hline
$\xi = 0.15$  & 34 & $71$   \\ \hline \hline
Dataset & \multicolumn{2}{|c|}{UTK} \\ \hline 
thresh.  & $\tau = 10^{-4}$ & $\tau = 10^{-5}$   \\ \hline 
$\xi = 0.05$  & 22 & 28   \\ \hline 
$\xi = 0.15$  & 25 & 31   \\ \hline 
\end{tabular}
\label{table:noise}
\end{center}
\end{table}

When estimating $R_c$ we used $p(\pi|i)$ and $p(\pi|b)$, which are related to predicting pixel $\pi$ from image $i$ and basis image $b$. Formally, we can estimate $R_c$ also via $p(i|\pi)$ and $p(i|b)$, since (\ref{a5}) holds for them as well; see the proof after (\ref{a5}). Such an estimation does not seem meaningful, because 
it compares prediction of image $i$ from quite different objects: pixel $\pi$ and basis image $b$. We implemented (\ref{a5}) for $p(i|\pi)$ and $p(i|b)$ and found consistently larger values of $R_c$ than those in Table~\ref{table:1}. This is expected for the less meaningful comparison.

\begin{figure}[th]
    \centering
    \includegraphics[width=0.99\linewidth]{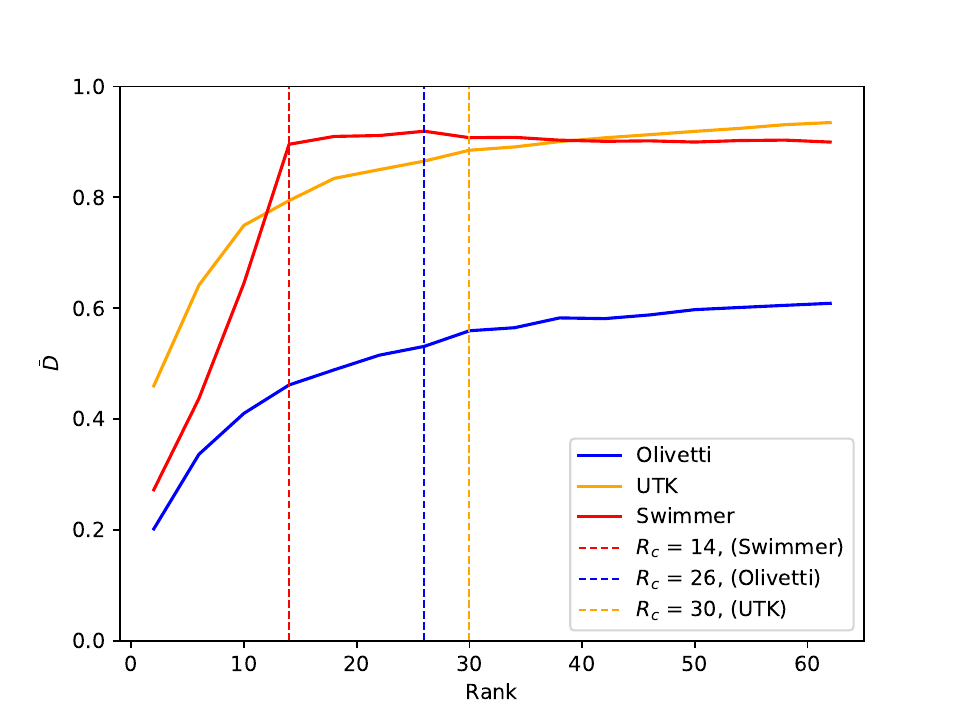}
    % \subfigure[$\bar{D}$ for Swimmer.]{

    %     \label{fig:swim_meanD.}
    % }
    % % \hspace{0.05\linewidth}
    % \subfigure[$\bar{D}$ for Olivetti and UTKFace.]{
    %     \includegraphics[width=0.45\linewidth]{Images_paper/oliv+utk150_10averaged_2-63-4.png}
    %     \label{fig:utkoliv_meanD.}
    % }
    \caption{ The average distance $\bar{D}$ between basis images [see (\ref{bard})] {\it vs.} the rank $R$ for three datasets: Swimmer, Olivetti and UTK; see section \ref{predo}. For each rank, $\bar{D}$ is evaluated for 10 different seeds and the mean is taken. For $\bar D({\rm UTK})$ we employed a random sample of 1500 images from UTK. For each dataset we show the estimated $R_c$ for clean datasets; see Table~\ref{table:1}. }
    \label{fig:meanD.}
\end{figure}

\subsection{Mean internal distance}

Here is an interesting structural aspect of NMF, which so far was not noticed in the literature, but which is related to $R_c$. Recall the cosine distance between vectors $v_1$ and $v_2$:
\be
D(v_1,v_2)=
1-v_1\cdot v_2\,\Big[(v_1\cdot v_1)\, (v_2\cdot v_2)\Big]^{-1/2},
\label{cos}
\ee
where $D(0,v)=D(v,0)=1$. Now define for (\ref{a1}) the mean cosine distance between basis images:
\be
\bar{D}=\frac{2}{R(R -1)}{\sum}_{a < b }^{R}D(B_{a}, B_{b}), ~~~ B_{a}=\{B_{\pi a}\}_{\pi=1}^N.
\label{bard}
\ee
$D(B_{a}, B_{b})$ is invariant against (\ref{a2}). $\bar{D}$ increases with $R$ and saturates at certain values of $R$; see Fig.~\ref{fig:meanD.}. The saturated value of $\bar{D}$ is larger than the maximal cosine distance between (original) images. This maximal distance for Swimmer, Olivetti and UTK equals (resp.) 0.421, 0.20548, and 0.6384; cf. saturated values in Fig.~\ref{fig:meanD.}. This is consistent with the fact that (for the considered $R$) basis images are sparser than images; see (\ref{oroihon}). The saturated value of $\bar{D}$ for Swimmer and UTK are comparable with each other and larger than the saturated value for Olivetti; see Fig.~\ref{fig:meanD.}. This is natural because Olivetti also contains images of the same person; see section \ref{dato}. Now, for clean datasets, there is a relation between $R_c$ and $\bar D$: $R_c$ is located at the edge of the saturation of $\bar D$; see Fig.~\ref{fig:meanD.}. This is especially clear for Swimmer. Elsewhere, we shall explore more relations between $R_c$ and $\bar D$.

\subsection{Comparison with Bayesian Information Criteria}
\label{bayo}

The problem of determining the effective rank $R$ for NMF is well-known and attracted much attention; see \cite{fogel,mdl,gillis-review} for reviews. Hence, we shall compare the above results with those found via the Bayesian Information Criteria (BIC); see \cite{bic_review, bic_derivation} for reviews. BIC emerged from within Bayesian probability theory, where a good measure of the weight of evidence for data $d$ to support a hypothesis $h$ (against its complement $\bar h$) was found to be $\ln[p(d|h)/p(d|\bar h)]$, where $p(d|h)$ is the conditional probability \cite{good_evidence}. Later, it was generalized to the limit of a large number of parameters, which led to the asymptotic relation $\ln p(d|h)\propto-\frac{1}{2}{\rm BIC} $. For NMF, BIC depends on $N$, $M$, the optimal error found from (\ref{fro}), and the parameter $R$ in (\ref{a1}) \cite{fogel}. Formulas for BICs are discussed in Ref.~\cite{fogel}; see Eqs.~(8-10) there. 

\begin{figure*}[t]
    \centering
    \includegraphics[width=1.0\linewidth]
    {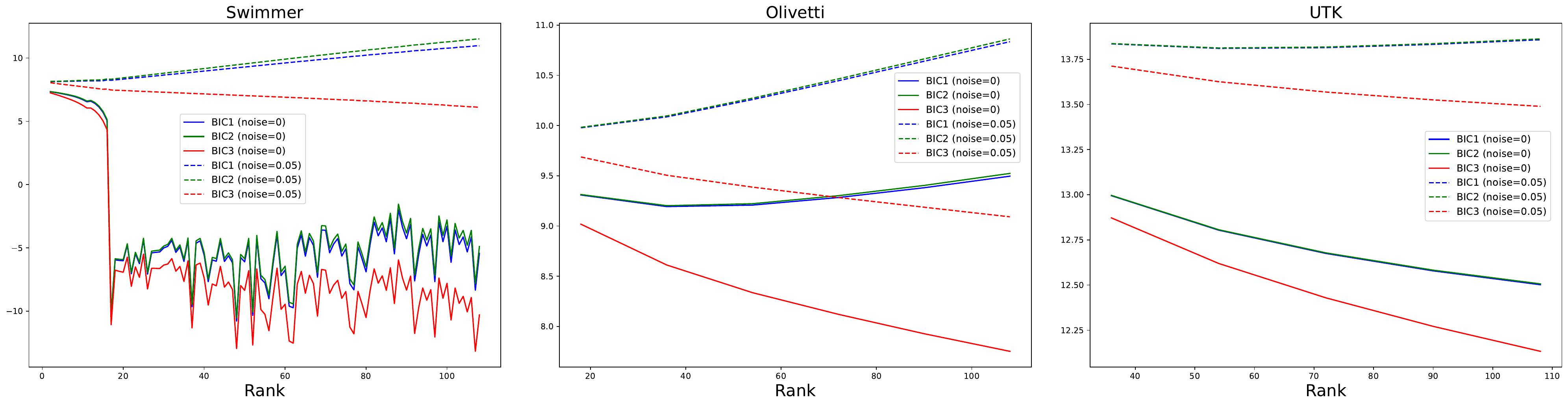}
    \caption{For clean and noisy datasets (Swimmer, Olivetti, UTK) we implemented 3 versions of BIC (Bayesian Information Criterion) that are defined and discussed in Ref.~\cite{fogel}; see section \ref{bayo}. The noise probability is given by (\ref{noise}). 
     First figure (Swimmer): for the clean case, the three BICs have a local minimum at $R=17$, which is the correct prediction, since it coincides with the nonnegative rank; cf.~the discussion in section \ref{data}. For a weak noise, none of the BICs show a local minimum, which is a nonsense behavior. Second figure (Olivetti): for the clean case, BIC1 and BIC2 are minimized at a reasonable value $R\approx 37$. For a weak noise they do not minimize. BIC3 is nonsensical. Third figure (UTK): for the clean case, none of the BICs are sensible. For a weak noise, BIC1, BIC2, and BIC3 do not show local minima.  
     }
    \label{fig:BIC}
\end{figure*}

We implemented BIC to determine the effective rank in (\ref{a1}, \ref{a3}) for the three clean and noisy datasets; see Fig.~\ref{fig:BIC}. The noise was added via (\ref{noise}) for $\xi=0.05$ (weak noise). Note that such BICs come in three versions: BIC1, BIC2, and BIC3 \cite{fogel}. Fig.~\ref{fig:BIC} shows that BICs can sometimes reproduce correct results; e.g. for the clean Swimmer dataset, the three versions of BIC are locally minimized around $R=17-18$, i.e., they agree with each other and do reproduce the real positive rank ${\rm rank}_+[P]=17$ of the clean Swimmer; cf.~the discussion in section \ref{data}. Altogether, the behavior of BICs is not sensible. They are not minimized in the weakly noisy Swimmer and Olivetti, as well as for the clean UTK; see Fig.~\ref{fig:BIC}. 
We also checked an alternative measure, RRSSQ (relative root of sum of square differences), which is mostly the rescaled version of (\ref{fropo}). Its minimization is used as a selection criterion for $R$ \cite{fogel}. RRSSQ also does not have a minimum for weak noise. 

%In contrast to this, our estimate of $R$ is stable against weak noise $\xi=0.05$; see our discussion in section %\ref{Restimation}. 

\section{Approximate PCC}
\label{appro}

We return to the problem {\it (iv)} in the introduction. We want to understand what type of approximation is implemented in (\ref{a3}) for $R<{\rm rank}_+[\{p(\pi,i)\}]$. The approximation concerns the joint probability $p(\pi,i)$, but does not concern the marginal probabilities $p(\pi)$ and $p(i)$, which are reproduced precisely according to (\ref{norma}). 

\subsection{Which joint probabilities are explained with a better approximation? }

To understand the approximation involved in (\ref{a3}), define
\begin{align}
\label{epso}
&\epsilon=\Big\{{|p(\pi|i)-\hat p(\pi|i)|}\Big/{p(\pi|i)}\Big\}_{\pi=1\,i=1}^{NM},\\
&w=\{{p(\pi|i)}\}_{\pi=1\,i=1}^{NM},
\label{epso2}
\end{align}
where $\epsilon$ is the sequence of relative errors, and where we define 
$\epsilon_{\pi i}=0$ for $p(\pi|i)=0$. According to (\ref{norma}), (\ref{epso}) can be also written as $\epsilon=\{{|p(\pi,i)-\hat p(\pi,i)|}/{p(\pi,i)}\}_{\pi=1\,i=1}^{NM}$. Treat $\epsilon$ and $w$ as $NM$-dimensional vectors, and note from (\ref{cos}) that $1-D(\epsilon-\bar\epsilon,w-\bar w)$ is the Pearson correlation coefficient between $\epsilon$ and $w$, where $\bar\epsilon=\frac{1}{NM}\sum_{\pi, i=1}^{NM}\epsilon_{\pi i}$ and $\bar w=1/N$ are the empiric means. Table~\ref{table:pearson} shows that within (\ref{a3}) there is anticorrelation 
\be
1-D(\epsilon-\bar\epsilon,w-\bar w)<0,
\label{rabat}
\ee
both according to Frobenius distance minimization and
KL-divergence minimization; see (\ref{fropo}) and (\ref{ku}).
These anticorrelations are weak, but, due to 
the large value of the sequence length $NM$, they are statistically significant. Eq.~(\ref{rabat}) means that in (\ref{a3}), larger probabilities $p(\pi|i)>\frac{1}{N}$ are approximated better than the smaller probabilities. Similar results (as in Table~\ref{table:pearson}) are obtained if the second sequence in (\ref{rabat}) is replaced by $\{{p(\pi,i)}\}_{\pi=1\,i=1}^{NM}$. This sequence differs from $\{{p(\pi|i)}\}_{\pi=1\,i=1}^{NM}$ by $p(i)=p(\pi,i)/p(\pi|i)$, which turns out to be a sufficiently homogeneous sequence so as not to significantly affect Pearson's coefficient.

\comment{A rough intuition for this anticorrelation is seen from looking at the minimized Frobenius metrics: $\sum_{\pi i=1}^{NM} (p(\pi,i)-\hat p(\pi,i))^2$. The anticorrelation follows if we assume that each factor inside the sum has the same order of magnitude.}

\begin{table}[t]
\caption{The Pearson correlation between the sequences (\ref{epso}) and (\ref{epso2}) at different ranks $R$. We considered Olivetti and UTK datasets. For UTK we worked with a random sample of 1500 images (the original size of UTK is 20000 images). ``Frobenius'' and ``KL'' indicate that the local minimization for NMF was achieved via (resp.) (\ref{fropo}) and (\ref{ku}).}
\centering
\begin{tabular}{|c||c|c|c|c| }
\hline
Dataset &  \multicolumn{2}{|c|}{Olivetti} & \multicolumn{2}{|c|}{UTK} \\ 
\hline
loss  & Frobenius & KL & Frobenius & KL \\ \hline 
$R=25$  & -0.3768 & -0.3679 & -0.2332 & -0.2247 \\ \hline 
$R=35$  & -0.3728 & -0.3518 & -0.2291  & -0.2192 \\ \hline 
$R=45$  & -0.3728 & -0.3524 & -0.2274  & -0.2159 \\ \hline 
$R=55$  & -0.3705 & -0.3415 & -0.2235 & -0.2102 \\ \hline 
$R=65$  & -0.3604 & -0.3509 & -0.2191 &  -0.2112 \\ \hline 
\end{tabular}
\label{table:pearson}
\end{table}

\begin{table}[t]
\caption{The Pearson correlation between the two sequences in (\ref{v}). Other parameters and notations are those of
Table \ref{table:pearson}.}
\centering
\begin{tabular}{|c||c|c|c|c| }
\hline
Dataset &  \multicolumn{2}{|c|}{Olivetti} & \multicolumn{2}{|c|}{UTK} \\ 
\hline
loss  & Frobenius & KL & Frobenius & KL \\ \hline 
$R=25$  & -0.3329 & -0.3229 & -0.2255 & -0.2166 \\ \hline 
$R=35$  & -0.3246 & -0.3053 & -0.2209  & -0.2108 \\ \hline 
$R=45$  & -0.3256 & -0.3025 & -0.2186  & -0.2075 \\ \hline 
$R=55$  & -0.3202 & -0.2880 & -0.2157 & -0.2022 \\ \hline 
$R=65$  & -0.3099 & -0.2940 & -0.2110 &  -0.2022 \\ \hline 
\end{tabular}
\label{table:pearson23}
\end{table}

\begin{figure}[t]
    \centering
    \includegraphics[width=0.99\linewidth]{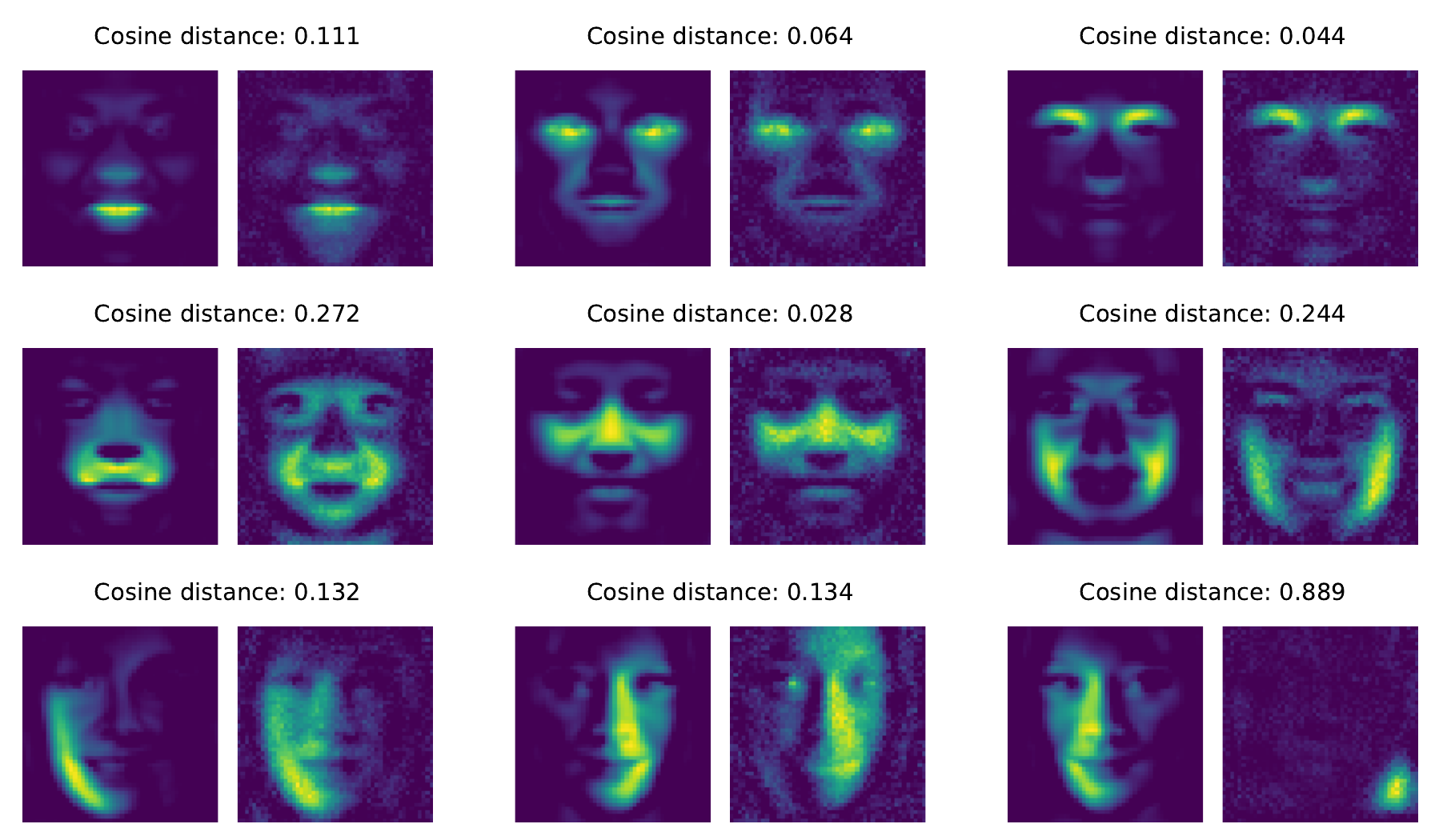}
    \caption{For each pair of pictures, the left image is a basis image from applying NMF (\ref{a1}, \ref{fro}, \ref{fropo}) with $R=36$ to the first half of UTK dataset. The second image of each pair is found in the same way, but beforehand we add to each image of the second half the noise following (\ref{noise}) with $\xi=0.25$. Both halves were trained with different seeds. 
    Basis images were optimally matched into pairs following the discussion around (\ref{kuhn}). Cosine distance means distance between a given image and its optimal match.
    It is seen that certain images which are not close to their match by distance, are nevertheless close semantically. For the considered 36 pairs, the mean cosine distance and the median are (resp.) 0.0236 and 0.0615.  
        }
\label{fig:UTK dataset metaimage matching examples.}
\end{figure}

\begin{figure*}[t]
    \centering
    \includegraphics[width=1.0\linewidth]{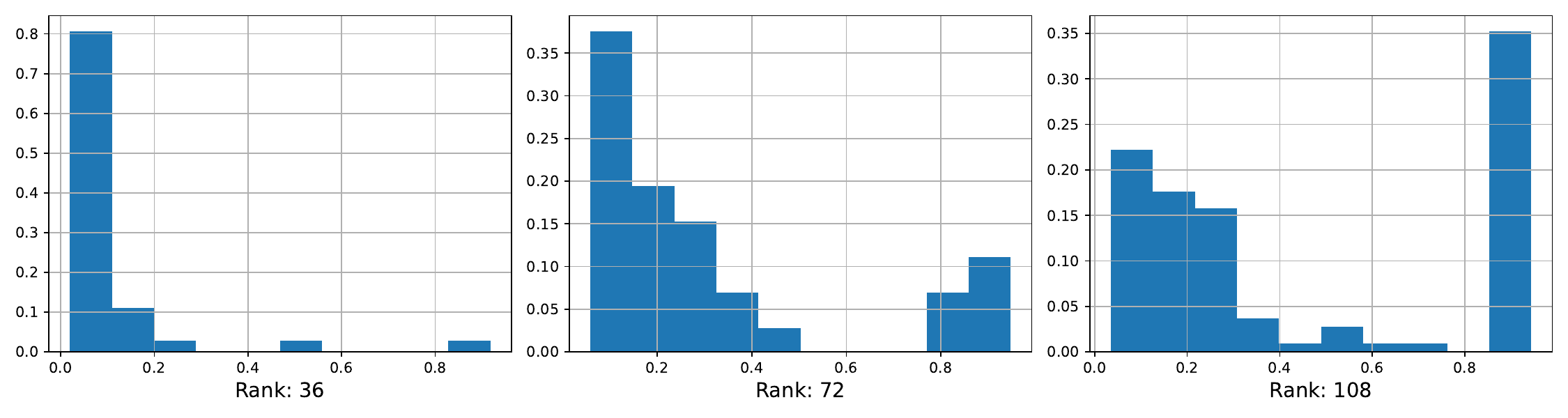}
    \caption{ 
The same situation as in Fig.~\ref{fig:UTK dataset metaimage matching examples.} (see the discussion around (\ref{kuhn})), but now we show the histograms of the distribution of optimally matched distances 
$\{ D_{a\,\hat{a}(a)}\}_{a=1}^R$ [see (\ref{cos}, \ref{kuhn}) between the two sets of basis images (extracted from the original and noise-corrupted halves of UTK dataset) for different values of the effective rank $R$. 
Note that for each rank, the halves are trained with the same seed. For each subfigure, the $x$-axis shows the binned values of $D_{a\,\hat a(a)}$, while the $y$-axis shows the percentages. For each rank $R$ the minimum, maximum, mean, and median of $\{ D_{a\,\hat{a}(a)}\}_{a=1}^R$
can be found in Table~\ref{table:R}. It is seen that as the effective NMF rank increases, basis images become less stable. 
        }
    \label{fig:UTK dataset metaimage matching distance distributions.}
\end{figure*}

\begin{figure*}[t]
    \centering
    \includegraphics[width=1.0\linewidth]{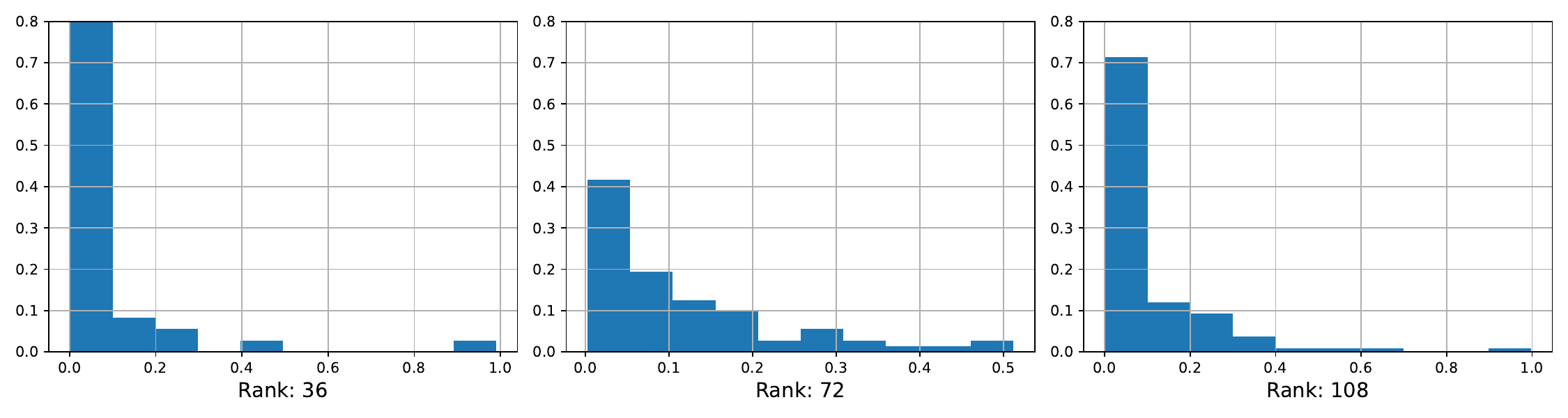}
    \caption{ 
The same situation as in Fig.~\ref{fig:UTK dataset metaimage matching distance distributions.}, but now noise is absent, and we compare (for each rank) the basis images generated via two different seeds applied to the same UTK data. The distance distribution becomes wider upon increasing $R$ from $R=36$ to $R=72$, and gets narrower when $R$ changes from $R=72$ to $R=108$; see Table~\ref{table:R} for means and medians.
        }
    \label{fig:UTK dataset metaimage matching distance distributions seeds.}
\end{figure*}

We observed another anticorrelation (see Table \ref{table:pearson23}):
\be
\label{v}
1-D(\epsilon-\bar\epsilon,v)<0,\quad
v=\{{p(\pi|i)-p(\pi)}\}_{\pi=1\,i=1}^{NM},
\ee
where $\bar v=0$. The meaning of (\ref{v}) is that the probabilities holding $p(\pi|i)>p(\pi)$ (positive correlation between $\pi$ and $i$) satisfy (\ref{a3}) better. Now $p(\pi|i)>p(\pi)$ has a certain causal meaning: $i$ increases the probability of $\pi$. This is a long-sought aspect of PCC \cite{uffink}: Reichenbach \cite{reich} proposed that only such probabilities $p(\pi|i)>p(\pi)$ are to be explained via conditional independence, but his implementation of this condition was formal, because the exact (\ref{a33}) explains all probabilities \cite{uffink}. 

In view of (\ref{rabat}, \ref{v}), NMF generalizes the exact PCC (\ref{a33}): probable and positively correlated events are better explained via conditional independence; cf.~section \ref{commi}. 

\subsection{Entropic relations}
\label{entrop}

NMF and PCC relate to interesting entropic relations. The first of them states that the approximate PCC increases entropy at the level of a single image:
\begin{align}
\label{ss}
& S_i {\leq }\hat{S}_i~~{\rm for~~all}~~i=1,...,M,\\
&S_i=-{\sum}_{\pi=1 }^{N}p(\pi|i)\ln p(\pi|i),\,
\hat S_i=-{\sum}_{\pi=1 }^{N}\hat p(\pi|i)\ln \hat p(\pi|i).\nonumber
\end{align}
Note from (\ref{a3}) that for $R=1$ the joint probability is approximated by the product of its marginal probabilities: $\hat p(\pi,i)=p(\pi)p(i)$; i.e., (\ref{ss}) is non-trivial even for $R=1$. 

The second relation states that for $R\geq R_0$ basis images are (in average) sparser than images:
\BEA
%S(\Pi|{\cal I})\equiv 
{\sum}_{i=1}^M p(i)S_i>
{\sum}_{b=1}^R p(b) S(\Pi|b),
%\equiv S(\Pi|{\cal B})
\label{oroihon}
\EEA
where $S(\Pi|b)=-{\sum}_{\pi=1}^Np(\pi|b)\ln[{p(\pi|b)}]$ is the entropy of the basis image $b$. Sparser $b$ means smaller $S(\Pi|b)$. Likewise, $S_i$ quantifies the sparsity of image $i$. We checked that the conclusion of (\ref{oroihon}) on sparsity is consistent with other measures of sparsity, e.g., with Hoyer's sparsity measure. For Swimmer, Olivetti and UTK, $R_0$ equals (resp.) 10, 2 and 2.

\comment{
$\hat{S}_i$ tends to decrease with increasing $R$; e.g. for UTK, 
$\hat S_i(R=36)>\hat S_i(R=54)$ for 97 \% of images $i$, while
$\hat S_i(R=54)>\hat S_i(R=72)$ for 95.5 \% of images $i$.
This trend is understandable, since with larger $R$ the approximation in (\ref{a3}) becomes more precise.  
\be
-{\sum}_{\pi=1\,i=1 }^{NM}p(\pi,i)\ln p(\pi|i) \leq -{\sum}_{\pi=1 }^{N}p(\pi)\ln p(\pi). 
\label{star}
\ee
}

\section{Basis image stability}
\label{basistability}

It is known that the basis images found via (\ref{a1}) are semantically interpretable \cite{lee,gillis-review} specifically due to the positivity constraints in (\ref{a1}). Such an interpretability is not present in other matrix factorization methods; e.g. it is absent in PCA, because the factors deduced from PCA are not positive and hence do not refer to any image or a weight matrix. 
We witness this interpretability in Fig.~\ref{fig:UTK dataset metaimage matching examples.} for UTK basis images, since they highlight various parts of human faces. For basis images to be really useful for interpretation, they must be stable against noise and the seed of the local optimization in (\ref{a1}).

Aside of its practical meaning, the basis image stability will provide an effective solution for the 
nonidentifiability problem that plagues both NMF and PCC; see section \ref{nonident}.

To investigate the stability of the UTK basis images, we divide UTK into two halves, apply noise (\ref{noise}) to the second half, and run NMF (\ref{a1}, \ref{fro}, \ref{fropo}) with the same rank $R$ on two separate halves of the UTK dataset. The halves were trained with the same random seed, so as to isolate the effect of changing $R$. Since the UTK dataset is curated to avoid duplicate photos or multiple images of the same person [see section \ref{dato}], its halves can be treated as independent datasets, despite sharing some higher-level characteristics, which we expect to be reflected in basis images. Thus, we get two sets of basis images:
$\{B_{ b}\}_{b=1}^R$ and $\{B_{a}^{\rm [noisy]}\}_{a=1}^R$, where $B_{a}=\{B_{\pi a}\}_{\pi=1}^N$.
They are found from (\ref{a1}) and
\be
{P}_{\pi i}^{\rm [noisy]} \approx \hat{P}_{\pi i}^{\rm [noisy]}
\equiv {\sum}_{b=1}^{{R}} B^{\rm [noisy]}_{\pi b} W^{\rm [noisy]}_{bi}.
\label{a1noisy}
\ee
We calculate the cosine distance (\ref{cos}) between them: $D_{ab}= D(B_{a},B^{\rm [noisy]}_{b})$.
Now we need to match the elements of these two sets using $D_{ab}$. The matching was achieved via the Kuhn-Munkres algorithm (implemented via the \texttt{linear\_sum\_assignment} function in SciPy \cite{scipy})
that addresses the \textit{linear assignment problem}:
\begin{align}
\label{kuhn}
\min_{X} \Big[{\sum}_{a,b}^R D_{ab} X_{ab}\,\Big|\,
{\sum}_{a=1}^R X_{ab} ={\sum}_{b=1}^R X_{ab} = 1\Big],
\end{align}
where $ a,b \in \{1, \dots, R\}$, and 
where \( X_{ab} = \{0, 1\} \) encodes the assignment (1 = assigned, 0 = unassigned). 
According to (\ref{kuhn}), matches the basis images with each other
by looking at the minimal sum of resulting cosine distances. The distance vector between the optimal matches reads from (\ref{kuhn}): $\{ D_{a\,\hat{a}(a)}\}_{a=1}^R$, where $\hat a(a)$ is the optimal match of $a$. 

Fig.~\ref{fig:UTK dataset metaimage matching examples.} illustrates matching results for $R=36$ and noise  $\xi=0.25$ [see (\ref{noise})] by presenting examples of close and far away basis images (found via the same seed). There are a few far away basis images at $R=36$: the mean, median, maximum and minimum of $\{ D_{a\,\hat{a}(a)}\}_{a=1}^{36}$ can be found at Table~\ref{table:R}; see also
Fig.~\ref{fig:UTK dataset metaimage matching distance distributions.}. Thus, at $R=36$ the basis images are stable with respect to a sizable noise $\xi=0.25$. We emphasize that $R=36$ for UTK is close to predictability estimate of the effective rank; see Table~\ref{table:1} and recall from section \ref{predo} that for UTK $R_c(\xi=0)=30$ and $R_c(\xi=0.25)=46$.

Fig.~\ref{fig:UTK dataset metaimage matching distance distributions.} shows the distribution of 
$\{ C_{a\,\hat{a}(a)}\}_{a=1}^R$ for various values of $R$. The stability is gradually lost once $R$ changes to $R=72$. For $R=108$, the number of close and far away basis images is comparable to each other. 

Fig.~\ref{fig:UTK dataset metaimage matching distance distributions seeds.} shows the distribution of basis image distances, when no noise is present, but the local optimization in (\ref{a1}) was initiated with different seeds applied to the same UTK data. (The distances are defined as in (\ref{a1noisy}, \ref{kuhn}).) Thereby, we learn about the local version of the nonidentifiability issue discussed in section \ref{nonident}. Indeed, since Fig.~\ref{fig:UTK dataset metaimage matching distance distributions seeds.} refers to the same data, the distribution of inter basis image distance is due to different local minima generated by different seeds. Fig.~\ref{fig:UTK dataset metaimage matching distance distributions seeds.} demonstrates that this nonidentifiability shows up for $R\ll {\rm rank}_+[P]$; see Table~\ref{table:R}. 
Fig.~\ref{fig:UTK dataset metaimage matching distance distributions seeds.} also shows that the behavior of the distance distribution is not monotonic upon changing $R$: $R=36$ and $R=108$ are similar to each other and differ from $R=72$. It is not yet clear why this occurs or whether it is a general phenomenon.

\begin{table}[htbp]
\caption{Parameters of distance distributions in (resp.)
Figs.~\ref{fig:UTK dataset metaimage matching distance distributions seeds.} (seed difference)
and \ref{fig:UTK dataset metaimage matching distance distributions.} (noise difference).
}
\begin{center}
\begin{tabular}{|c||c|c|c|c| }
\hline
       & {\rm mean} & {\rm median} & {\rm max} & {\rm min} \\ 
\hline
Fig.~ \ref{fig:UTK dataset metaimage matching distance distributions.}: $R=36$
       &0.1056 &0.0638 &0.9175 &0.0195 \\ 
       \hline
Fig.~ \ref{fig:UTK dataset metaimage matching distance distributions.}:~$R=72$~ 
& 0.3106 & 0.1958 & 0.9481 & 0.0584 \\
       \hline       
~Fig.~ \ref{fig:UTK dataset metaimage matching distance distributions.}:
$R=108$ & 0.456 & 0.2752 & 0.944 & 0.0348  \\ 
       \hline       
       \hline
Fig.~\ref{fig:UTK dataset metaimage matching distance distributions seeds.}: $R=36$
       & 0.0857 & 0.0171 & 0.9901 & 0.0012\\ \hline
Fig.~\ref{fig:UTK dataset metaimage matching distance distributions seeds.}: $R=72$ 
& 0.1151
& 0.0806
& 0.5123
& 0.0024
\\ 
       \hline       
~Fig.~\ref{fig:UTK dataset metaimage matching distance distributions seeds.}: $R=108$ 
& 0.0951
& 0.0357
& 0.9977
& 0.0009
\\ 
       \hline       
\end{tabular}
\label{table:R}
\end{center}
\end{table}

\section{Natural clusterization of images}
\label{nato}

\begin{figure}[th]
    \centering
    \includegraphics[width=0.99\linewidth]{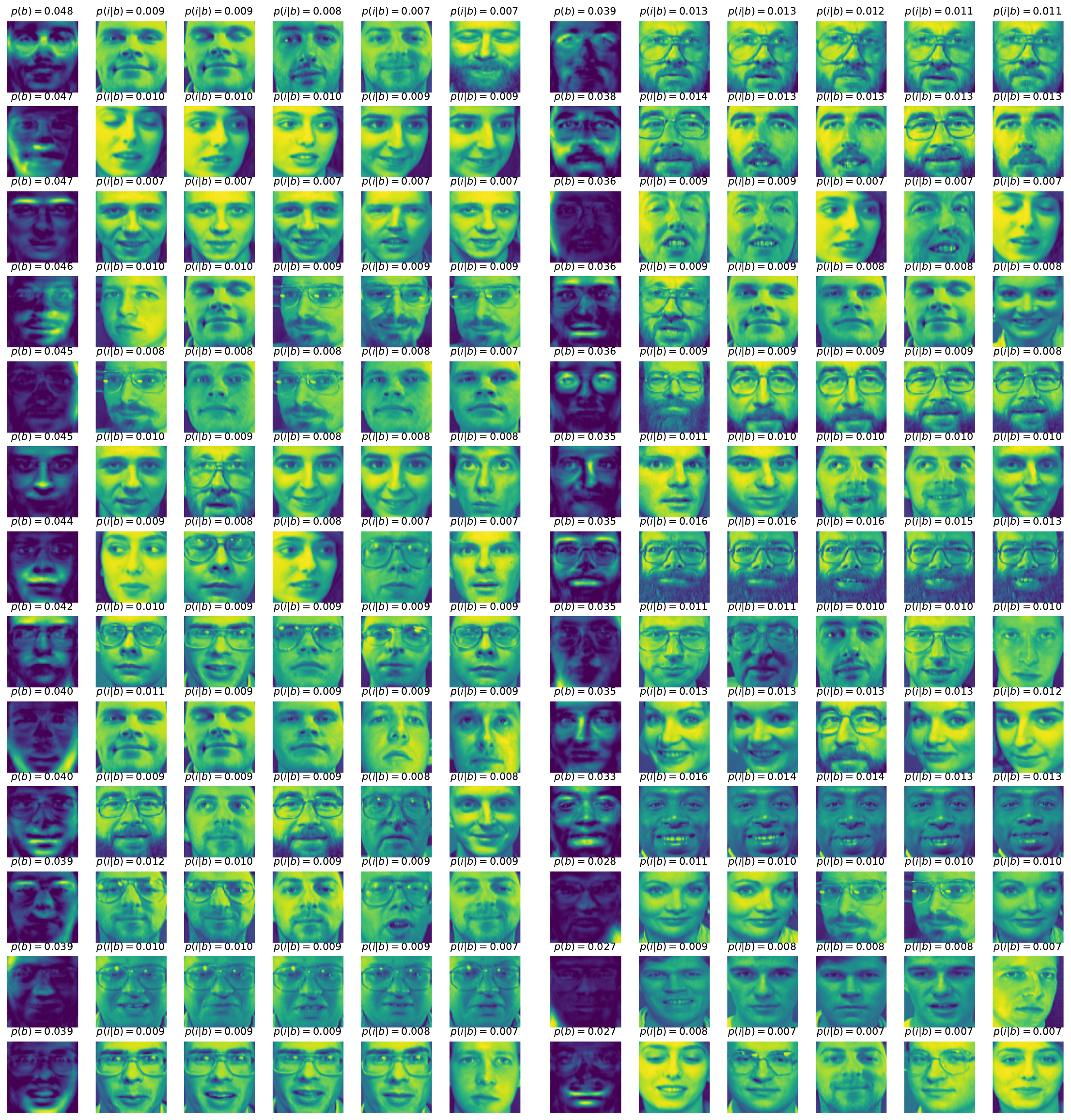}
    \caption{Visualization of natural clusterization via NMF for Olivetti dataset with $R=26$. 
    In each group of 26 images (arranged in rows), there is a basis image as well as five cluster images derived from the basis image; see section \ref{nato}. Basis images $b$ are arranged with decreasing $p(b)$; cf.~(\ref{a303}). After zooming on a computer screen the reader will see the values of $p(i|b)$ and $p(b)$ (resp.) for every image and basis image. }
    \label{fig:cluster.}
\end{figure}

The idea of natural clusterization is well-accepted in biology \cite{sober_common}: organisms having the same ancestor (i.e. common cause) are put into the same cluster. Cognitive psychology recently developed a similar idea for explaining human classification behavior: objects having the same cause are put into the same class \cite{rehder2003causal}. This section aims to develop a similar idea in the context of NMF (\ref{a3}). It amounts to defining a class of images caused by a given basis image. 

Recall that $p(i|b)$ in (\ref{a3}, \ref{a303}) is the conditional probability of image $i$ given basis image $b$. A sufficiently large $p(i|b)-p(i)>0$ means a positive correlation between $i$ and $b$, and can be also interpreted as $b$ ``causing'' $i$ \cite{reich,suppes}. We define a class characterized (or driven) by the basis image $b$ by putting there all images with sufficiently large $p(i|b)-p(i)>0$. We implemented this for Olivetti dataset with $R=26$; cf.~Fig.~\ref{fig:meanD.}. For each (among $26$) basis image $b$ we selected 5 images $i$ that refer to the top largest values of $p(i|b)$ (condition $p(i|b)-p(i)>0$ holds for them). Fig.~\ref{fig:cluster.} shows the result, where each cluster indeed contains similar images. Frequently, these are images of the same person (Olivetti has this possibility). The method also produces sensible results for UTK (not shown).

Related ideas of causal feature extraction in machine learning are reviewed in \cite{guyon_review,chalupka_2}. The difference with the present approach is that we do not work with already existing features \cite{guyon_review}, do not deduce those features from coarse-graining \cite{chalupka_2}, and do not assume directed acyclic graphs and do-calculus \cite{chalupka_2}. In contrast, our features are deduced from (\ref{a1}). 

\section{Denoising}
\label{denoise}

Image denoising is a traditional task of machine learning, which generated many approaches; see \cite{review_denoising} for review. We focus on dictionary based methods for denoising, where the needed
prior information comes from the fact that the image belongs to a specific dataset \cite{review_denoising}. 
This task does not need interpretability in terms of positive weights and basis images. 
Such methods frequently employ matrix factorization methods, e.g., PCA. We aim to see to which extent NMF applies to the denoising problem, and compare it with PCA. We are going to elucidate the role of the effective rank in denoising. 

\begin{table}%[htbp]
\caption{$R_1$ and $R_2$ are defined in (\ref{alister}). UTK-1 and UTK-2 are (resp.) 1500 
and 3000 sample subsets of the full UTK dataset. 
The noise is generated by (\ref{noise}) with $\xi=0.25$. 
For Swimmer dataset we allow not more than 2 
exclusion for $M$ inequalities (\ref{alister}). 
}
\begin{center}
\begin{tabular}{|c||c|c|c|c| }
\hline
Dataset & Swimmer & Olivetti & UTK-1 & UTK-2 \\ \hline
$R_1$ & 1 & 2 & 7 & 7 \\ \hline
$R_2$ & 42 & $236$ & 134 & 154 \\ \hline
\end{tabular}
\label{table:r12}
\end{center}
\end{table}

\begin{figure}[t]
    \centering
    \includegraphics[width=1.1\linewidth]{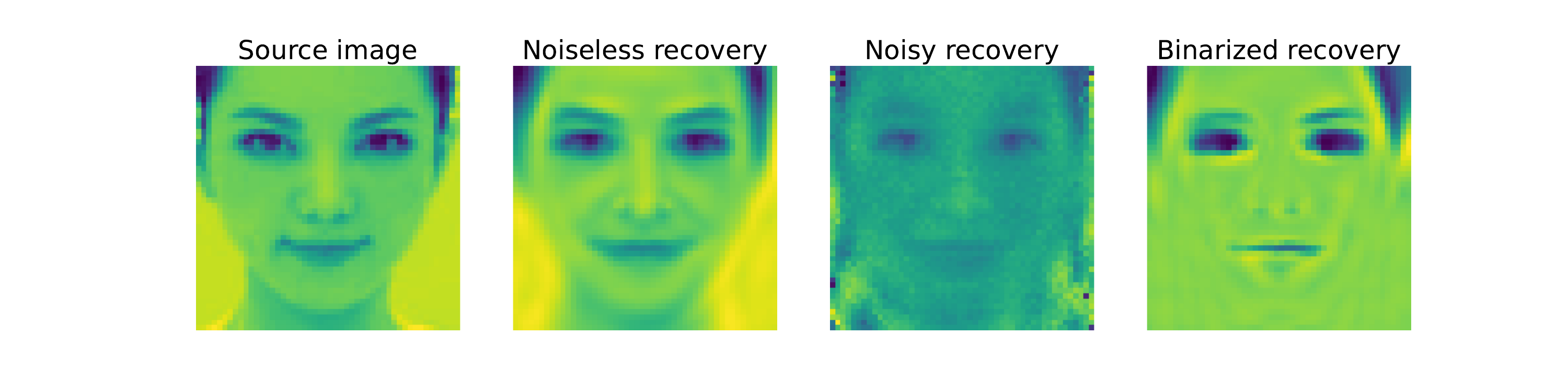}
    \caption{ Recovery of a UTK image via NMF with rank $R=108$. The first image is the original; the second one is found from NMF (\ref{a1}). The third one is found from the original image after subjecting it to noise (\ref{noise}) with $\xi=0.25$ and then recovering via (\ref{a1noisy}). The last image is similar to the third one, but instead of noise (\ref{noise}), we applied binarization; see section \ref{dato}.}
    \label{fig:UTK recovery examples.}
\end{figure}

Let us start from (\ref{a1}, \ref{a1noisy}). NMF denoises a given dataset, if for each image $i$ (or for a majority of them), $P_{i}=\{P_{\pi i}\}_{\pi=1}^N$ is closer to $\hat P_{i}^{\rm [noisy]}=\{\hat P_{\pi i}^{\rm [noisy]}\}_{\pi=1}^N$ than $P_{i}$ to $P_{i}^{\rm [noisy]}=\{P_{\pi i}^{\rm [noisy]}\}_{\pi=1}^N$. 
Using (\ref{cos}) we formalize this condition as
\be
\label{alister}
D(P_{i},  P_{i}^{\rm [noisy]})-D(P_{i}, \hat P_{i}^{\rm [noisy]})>0, ~~ i=1,...,M.
\ee
Eqs.~(\ref{alister}) are satisfied for $R_1\leq R\leq R_2$ in all datasets we studied; see Table~\ref{table:r12}. $R_1$ and $R_2$ depend on the noise and dataset. Fig.~\ref{fig:UTK recovery examples.} presents examples of $P_{i}$, $\hat P_{i}$ and $\hat P_{i}^{\rm [noisy]}$. For the strong noise with $\xi=0.25$, the recovery results in 
Fig.~\ref{fig:UTK recovery examples.} are satisfactory. 

\comment{
If the image distortion in UTK is produced by binarization [see section \ref{data}], (\ref{alister}) hold for 99 \% of images. }

Some inequalities in (\ref{alister}) are violated for $R<R_1$, since no reliable set of basis images is found. Others are violated for $R>R_2$, because $\{\hat P_{i}^{\rm [noisy]}\}_{i=1}^M$ overfit the noise. Note from Table~\ref{table:r12} that larger samples of UTK have larger prior information, and hence larger $R_2$. Likewise, Olivetti has many similar images, which increases its relative prior information and leads to a large $R_2$; see Table~\ref{table:r12}. 

We apply various measures to get more precise information on $R$ in the range $R_1\leq R\leq R_2$. For example, in UTK, $|P-\hat P^{\rm [noisy]} |_2$ is minimized for a certain $R_{\rm min}\in [R_1,R_2]$ illustrating the phenomenon of overfitting; cf.~(\ref{fropo}, \ref{a1noisy}). 
Another concrete quantity is the accuracy of denoising: define $\kappa_i=1$ if the original image $P_i$ is closest (in sense of (\ref{cos})) to $\hat P_{i}^{\rm [noisy]}$; otherwise, set $\kappa_i=0$. The accuracy (AC) reads
$\text{AC} = \frac{1}{M} {\sum}_{i=1}^{M} \kappa_i$. Larger AC means better denoising. Denoising can be also implemented via PCA, with the same definition of AC.

\begin{figure}[t]
    \centering
    \includegraphics[width=1.1\linewidth]{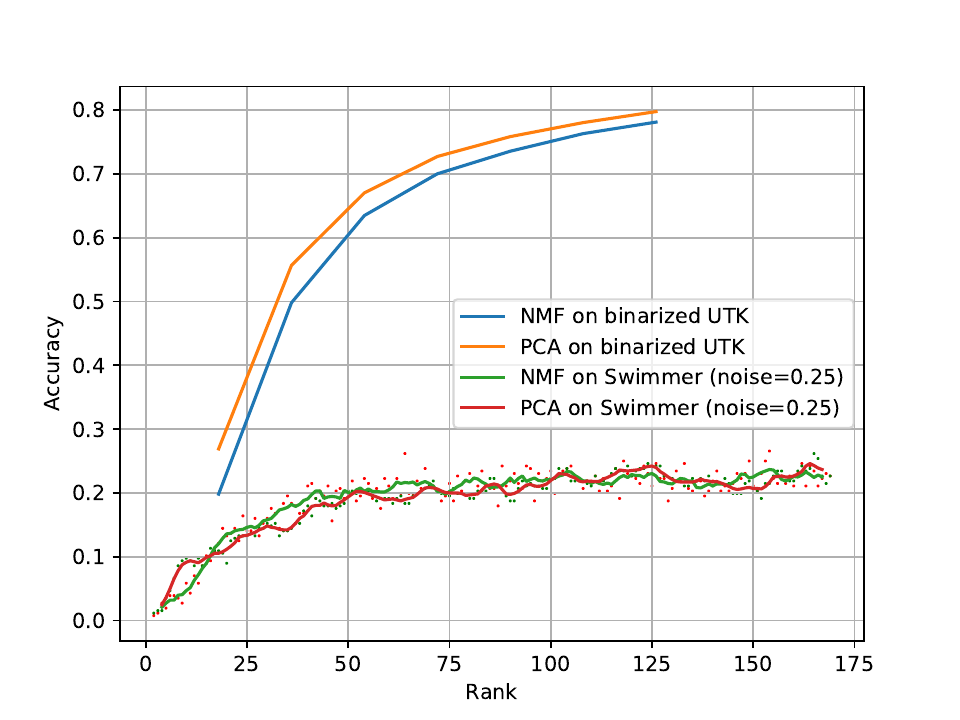}
    \caption{Denoising accuracy comparison for Swimmer and UTK datasets. Two main matrix factorization methods are compared: PCA and NMF. The noise probability for Swimmer dataset is $\xi=0.25$. Because of this, the accuracy score (green and red dots) shows a noisy dependence on rank, so we also plot the mean average graph with a window size of 5.
        }
\label{fig:denoising}        
\end{figure}

Fig.~\ref{fig:denoising} shows the denoising accuracy results for Swimmer dataset under a strong noise: (\ref{noise}) with $\xi=0.25$. We see that ${\rm AC}_{\rm NMF}>{\rm AC}_{\rm PCA}$ is possible. 
Fig.~\ref{fig:denoising} also shows the denoising accuracy results for UTK. Here the distortion is introduced via binarization; see section \ref{data}. Now ${\rm AC}_{\rm PCA}>{\rm AC}_{\rm NMF}$, and both accuracies grow monotonically with $R$. For both cases, AC is well above its random threshold $1/M$ obtained via a random permutation of $M$ images. Then, the probability of having $m$ images in their original places equals (for $M\gg 1$) $e^{-1}/m!$. Hence, the random AC is $\frac{1}{eM}\sum_m m/m!=\frac{1}{M}$. 

\comment{
We note that ${\rm AC}_{\rm NMF}$ and ${\rm AC}_{\rm PCA}$ saturate for $R>100$. This value of $R$ is larger than $R_c(\xi=0.25)=46$ for UTK; see section \ref{predo}. Denoising does not require interpretability, hence the optimal values of $R$ for this task need not be around $R_c$. For denoising UTK, the best accuracy is achieved at higher ranks than $R_c$, as Fig.~\ref{fig:denoising} shows. However, we know that increasing the rank reduces the stability of basis images; see Figs.~\ref{fig:UTK dataset metaimage matching distance distributions.} and
\ref{fig:UTK dataset metaimage matching distance distributions seeds.}. Thus, we have a tradeoff between interpretability and the quality of denoising. 
}

\section{Summary and open problems}

Open problems {\it (i)-(iv)} concerning Nonnegative Matrix Factorization (NMF) and the Principle of the Common Cause (PCC) were outlined in the introduction in the joint context of gray-scale images and probability models; see also section \ref{commi}. Here are our responses to them.

-- The effective rank $R$ of NMF was estimated via the predictability method deduced from the exact PCC; see section \ref{Restimation}. In contrast to estimates found via BIC (Bayesian Information Criteria), our estimate $R_c$ is stable against weak data noise. $R_c$ does have a geometrical meaning related to a limiting distribution of basis image distance, and it increases under strong noise; see section \ref{predo}. 

-- The advantage of NMF is the interpretability of its outcomes (basis images and weights). But to be useful the (interpretable) basis images have to be stable. Around $R_c$, basis images deduced from NMF are stable against noise and against changing the seed of the local optimization; see section \ref{basistability}. Both stabilities are lost for $R\gg R_c$; see Figs.~\ref{fig:UTK dataset metaimage matching distance distributions.} and \ref{fig:UTK dataset metaimage matching distance distributions seeds.}.

-- NMF applies to clustering: images having the same common cause basis image are put in the same cluster. This method produces reasonable results; see section \ref{nato}.

-- NMF can be applied to data denoising in a range $R_1\leq R\leq R_2$: for $R< R_1$, there is no denoising, since no reliable images are formed. For $R>R_2$ we have overfitting to noise; see section \ref{denoise}. Sometimes NMF denoising outperforms PCA in terms of accuracy. 

-- PCC is generalized in the context of NMF; see section \ref{appro}. In its original exact formulation, PCC makes conditionally independent all joint probabilities. In the approximate version inspired by NMF, 
only larger and positively correlated joint probabilities tend to be accurately explained.

There are several problems left for the future. {\it (1)} We identified several scenarios of nonidentifiability related to NMF; e.g. Fig.~\ref{fig:UTK dataset metaimage matching distance distributions seeds.} shows that different seeds applied to the same data lead to different local minima and (numerically) quite different basis images. This does not necessarily mean that these different basis images refer to different semantic features. To address this question, we are working on semantic analysis tools, which can apply to NMF. {\it (2)} Information theoretic tools can elucidate the structure of NMF and uncover its hidden features; see section \ref{entrop} for initial results. 
{\it (3)} NMF denoising needs further understanding in comparison to other matrix factorization methods.

\section*{Acknowledgment}

This work was supported by the HESC of Armenia under Grants 24FP-1F030 and 21AG-1C038. We thank H. Avetisyan, V. Bardakhchyan, H. Khachatryan, A. Matevosyan for support.

%\bibliography{paper}

\bibliographystyle{IEEEtran}
\bibliography{IEEEabrv,paper}

\begin{thebibliography}{10}
\providecommand{\url}[1]{#1}
\csname url@samestyle\endcsname
\providecommand{\newblock}{\relax}
\providecommand{\bibinfo}[2]{#2}
\providecommand{\BIBentrySTDinterwordspacing}{\spaceskip=0pt\relax}
\providecommand{\BIBentryALTinterwordstretchfactor}{4}
\providecommand{\BIBentryALTinterwordspacing}{\spaceskip=\fontdimen2\font plus
\BIBentryALTinterwordstretchfactor\fontdimen3\font minus \fontdimen4\font\relax}
\providecommand{\BIBforeignlanguage}[2]{{%
\expandafter\ifx\csname l@#1\endcsname\relax
\typeout{** WARNING: IEEEtran.bst: No hyphenation pattern has been}%
\typeout{** loaded for the language `#1'. Using the pattern for}%
\typeout{** the default language instead.}%
\else
\language=\csname l@#1\endcsname
\fi
#2}}
\providecommand{\BIBdecl}{\relax}
\BIBdecl

\bibitem{finn}
P.~Paatero and U.~Tapper, ``Positive matrix factorization: A non-negative factor model with optimal utilization of error estimates of data values,'' \emph{Environmetrics}, vol.~5, no.~2, pp. 111--126, 1994.

\bibitem{lee}
D.~D. Lee and H.~S. Seung, ``Learning the parts of objects by non-negative matrix factorization,'' \emph{Nature}, vol. 401, no. 6755, pp. 788--791, 1999.

\bibitem{gillis-review}
N.~Gillis, \emph{Nonnegative Matrix Factorization}.\hskip 1em plus 0.5em minus 0.4em\relax SIAM, 2021.

\bibitem{complexity-vava}
S.~A. Vavasis, ``On the complexity of nonnegative matrix factorization,'' \emph{SIAM journal on optimization}, vol.~20, no.~3, pp. 1364--1377, 2010.

\bibitem{fogel}
J.~M. Maisog, A.~T. DeMarco, K.~Devarajan, S.~Young, P.~Fogel, and G.~Luta, ``Assessing methods for evaluating the number of components in non-negative matrix factorization,'' \emph{Mathematics}, vol.~9, no.~22, p. 2840, 2021.

\bibitem{mdl}
S.~Squires, A.~Pr{\"u}gel-Bennett, and M.~Niranjan, ``Rank selection in nonnegative matrix factorization using minimum description length,'' \emph{Neural computation}, vol.~29, no.~8, pp. 2164--2176, 2017.

\bibitem{reich}
H.~Reichenbach, \emph{The direction of time}.\hskip 1em plus 0.5em minus 0.4em\relax University of California Press, 1956, vol.~65.

\bibitem{suppes}
P.~Suppes, \emph{A probabilistic theory of causality}.\hskip 1em plus 0.5em minus 0.4em\relax North-Holland, Amsterdam, 1970.

\bibitem{janzing}
J.~Peters, D.~Janzing, and B.~Sch{\"o}lkopf, \emph{Elements of causal inference: foundations and learning algorithms}.\hskip 1em plus 0.5em minus 0.4em\relax The MIT Press, 2017.

\bibitem{wharton}
K.~B. Wharton and N.~Argaman, ``Colloquium: Bell’s theorem and locally mediated reformulations of quantum mechanics,'' \emph{Reviews of Modern Physics}, vol.~92, no.~2, p. 021002, 2020.

\bibitem{penrose1962direction}
O.~Penrose and I.~C. Percival, ``The direction of time,'' \emph{Proceedings of the Physical Society}, vol.~79, no.~3, p. 605, 1962.

\bibitem{rehder2003causal}
B.~Rehder, ``A causal-model theory of conceptual representation and categorization.'' \emph{Journal of Experimental Psychology: Learning, Memory, and Cognition}, vol.~29, no.~6, p. 1141, 2003.

\bibitem{bio}
K.~Devarajan, ``Nonnegative matrix factorization: an analytical and interpretive tool in computational biology,'' \emph{PLoS computational biology}, vol.~4, no.~7, p. e1000029, 2008.

\bibitem{sober_common}
E.~Sober, ``Common cause explanation,'' \emph{Philosophy of Science}, vol.~51, no.~2, pp. 212--241, 1984.

\bibitem{simpson}
A.~Hovhannisyan and A.~Allahverdyan, ``Resolution of simpson's paradox via the common cause principle,'' \emph{arXiv preprint arXiv:2403.00957}, 2024.

\bibitem{review_denoising}
B.~Goyal, A.~Dogra, S.~Agrawal, B.~S. Sohi, and A.~Sharma, ``Image denoising review: From classical to state-of-the-art approaches,'' \emph{Information fusion}, vol.~55, pp. 220--244, 2020.

\bibitem{kl}
N.-D. Ho and P.~Van~Dooren, ``Non-negative matrix factorization with fixed row and column sums,'' \emph{Linear Algebra and its Applications}, vol. 429, no. 5-6, pp. 1020--1025, 2008.

\bibitem{scikit}
F.~Pedregosa, G.~Varoquaux, A.~Gramfort, V.~Michel, B.~Thirion, O.~Grisel, M.~Blondel, P.~Prettenhofer, R.~Weiss, V.~Dubourg, J.~Vanderplas, A.~Passos, D.~Cournapeau, M.~Brucher, M.~Perrot, and {\'E}.~Duchesnay, ``Scikit-learn: Machine learning in python,'' \emph{Journal of Machine Learning Research}, vol.~12, no.~85, pp. 2825--2830, 2011.

\bibitem{rank-nonnegative}
J.~E. Cohen and U.~G. Rothblum, ``Nonnegative ranks, decompositions, and factorizations of nonnegative matrices,'' \emph{Linear Algebra and its Applications}, vol. 190, pp. 149--168, 1993.

\bibitem{donoho}
D.~Donoho and V.~Stodden, ``When does non-negative matrix factorization give a correct decomposition into parts?'' \emph{Advances in neural information processing systems}, vol.~16, 2003.

\bibitem{hovh2023}
A.~Hovhannisyan and A.~E. Allahverdyan, ``The most likely common cause,'' \emph{International Journal of Approximate Reasoning}, vol. 173, p. 109264, 2024.

\bibitem{guyon_review}
I.~Guyon, C.~Aliferis \emph{et~al.}, ``Causal feature selection,'' in \emph{Computational methods of feature selection}.\hskip 1em plus 0.5em minus 0.4em\relax Chapman and Hall/CRC, 2007, pp. 79--102.

\bibitem{causal_non_causal_review}
K.~Yu, L.~Liu, and J.~Li, ``A unified view of causal and non-causal feature selection,'' \emph{ACM Transactions on Knowledge Discovery from Data (TKDD)}, vol.~15, no.~4, pp. 1--46, 2021.

\bibitem{mazzola2019generalised}
C.~Mazzola, ``Generalised reichenbachian common cause systems,'' \emph{Synthese}, vol. 196, no.~10, pp. 4185--4209, 2019.

\bibitem{horwich1987asymmetries}
P.~Horwich, \emph{Asymmetries in time: Problems in the philosophy of science}.\hskip 1em plus 0.5em minus 0.4em\relax MIT press, 1987.

\bibitem{uffink}
J.~Uffink, ``The principle of the common cause faces the bernstein paradox,'' \emph{Philosophy of Science}, vol.~66, no.~S3, pp. S512--S525, 1999.

\bibitem{imprecise}
D.~Kinney, ``Imprecise bayesian networks as causal models,'' \emph{Information}, vol.~9, no.~9, p. 211, 2018.

\bibitem{olivetti_faces_dataset}
{AT\&T Laboratories Cambridge}, ``The database of faces,'' \url{https://www.cl.cam.ac.uk/research/dtg/attarchive/facedatabase.html}, 1994.

\bibitem{zhifei2017cvpr}
Z.~Zhang, Y.~Song, and H.~Qi, ``Age progression/regression by conditional adversarial autoencoder,'' in \emph{IEEE Conference on Computer Vision and Pattern Recognition (CVPR)}.\hskip 1em plus 0.5em minus 0.4em\relax IEEE, 2017.

\bibitem{jangedoo_utkface}
Jangedoo, ``Utkface,'' \url{https://www.kaggle.com/datasets/jangedoo/utkface-new}.

\bibitem{bic_review}
P.~Stoica and Y.~Selen, ``Model-order selection: a review of information criterion rules,'' \emph{IEEE signal processing magazine}, vol.~21, no.~4, pp. 36--47, 2004.

\bibitem{bic_derivation}
H.~S. Bhat and N.~Kumar, ``On the derivation of the bayesian information criterion,'' \emph{School of Natural Sciences, University of California}, vol.~99, no.~4, 2010.

\bibitem{good_evidence}
I.~Good, ``Weight of evidence: A brief survey,'' \emph{Bayesian statistics}, vol.~2, pp. 249--270, 1985.

\bibitem{scipy}
P.~Virtanen, R.~Gommers, T.~E. Oliphant, M.~Haberland, T.~Reddy, D.~Cournapeau, E.~Burovski, P.~Peterson, W.~Weckesser, J.~Bright \emph{et~al.}, ``Scipy 1.0: Fundamental algorithms for scientific computing in python,'' \emph{Nature Methods}, vol.~17, no.~3, pp. 261--272, 2020.

\bibitem{chalupka_2}
K.~Chalupka, F.~Eberhardt, and P.~Perona, ``Causal feature learning: an overview,'' \emph{Behaviormetrika}, vol.~44, pp. 137--164, 2017.

\end{thebibliography}

\end{document}